\documentclass[authoryear,review,3p,times]{elsarticle}

\usepackage{amssymb}
\usepackage{amsmath}
\usepackage{lineno}
\usepackage{bm} 
\usepackage{longtable}
\usepackage{caption}
\usepackage{soul}
\usepackage{color}
\usepackage{gensymb}
\usepackage[flushleft]{threeparttable}
\usepackage{algorithm}
\usepackage{algpseudocode}
\usepackage{hyperref}
\usepackage{nomencl} 
\makenomenclature

\graphicspath{{figures/}}

\usepackage{array}

\journal{Nuclear Engineering and Design}

\begin{document}

\begin{frontmatter}



\title{Techno-economic optimization of a heat-pipe microreactor, part I: theory and cost optimization}


\author[1]{Paul Seurin \corref{cor1}} 
\ead{paul.seurin@inl.gov}
\author[2]{Dean Price} 
\author[1]{Luis Nunez} 

\affiliation[1]{organization={Autonomous Engineering Laboratory, Idaho National Laboratory}, 
            addressline={1955 Fremont Ave.}, 
            city={Idaho Falls},
            postcode={83402}, 
            state={ID},
            country={U.S.}}

\affiliation[2]{organization=Massachusetts Institute of Technology, 
            addressline={77 Massachusetts Ave.}, 
            city={Cambridge},
            postcode={02935}, 
            state={MA},
            country={U.S.}}

\begin{abstract}
Microreactors ($\mu$R)s, particularly heat-pipe microreactors (HPMRs), are compact, transportable, self-regulated power systems well-suited for access-challenged remote areas where costly fossil fuels dominate. However, they suffer from diseconomies of scale, and  their financial viability remains unconvincing. One step in addressing this shortcoming is to design these reactors with comprehensive economic and physics analyses informing early-stage design iteration. In this work, we present a novel unifying geometric design optimization approach that accounts for techno-economic considerations. We start by generating random samples to train surrogate models, including Gaussian processes (GPs) and multi-layer perceptrons (MLPs). We then deploy these surrogates within a reinforcement learning (RL)-based optimization framework to optimize the levelized cost of electricity (LCOE), all the while imposing constraints on the fuel lifetime, shutdown margin (SDM), peak heat flux, and rod-integrated peaking factor $F_{\Delta h}$. We study two cases: one in which the axial reflector cost is very high, and one in which it is inexpensive. We found that the operation and maintenance (O\&M) and capital costs are the primary contributors to the overall LCOE---particularly the cost of the axial reflectors (for the first case) and the control drum materials. The optimizer cleverly changes the design parameters so as to minimize one of them while still satisfying the constraints, ultimately reducing the LCOE by more than 57\% in both instances. A comprehensive integration of fuel and HP performance with multi-objective optimization is currently being pursued to fully understand the interaction between constraints and cost performance.
\end{abstract}




\end{frontmatter}

\nomenclature{QoI}{Quantities of interest}
\nomenclature{SDM}{Shutdown margin}
\nomenclature{$F_q$}{Pin peaking factor}
\nomenclature{$F_{\Delta h}$}{Rod-integrated peaking factor}
\nomenclature{$q^{''}_{max}$}{Peak heat flux}
\nomenclature{ITC}{Isothermal temperature coefficient}
\nomenclature{TC}{Temperature coefficient}
\nomenclature{CRD}{Control rod}

\nomenclature{RF}{Random forest}
\nomenclature{MLP}{Multi-layered perceptron}
\nomenclature{GP}{Gaussian process}
\nomenclature{AI}{Artificial intelligence}

\nomenclature{RL}{Reinforcement learning}
\nomenclature{PPO}{Proximal policy optimization}
\nomenclature{$\mu$R}{Microreactor}
\nomenclature{HPMR}{Heat-pipe microreactor}
\nomenclature{HP}{Heat pipe}
\nomenclature{TRISO}{Tri-structural isotropic particle}

\nomenclature{MOUSE}{Microreactor Optimization Using Simulation and Economics}
\nomenclature{LCOE}{Levelized cost of electricity}
\nomenclature{COA}{Code of account}
\nomenclature{O\&M}{Operation and maintenance}
\nomenclature{OCC}{Overnight capital cost}
\nomenclature{NOAK}{Nth of a kind}
\nomenclature{FOAK}{First of a kind}
\nomenclature{FTE}{Full-time equivalent}

\nomenclature{YHx}{Yttrium hydride}
\nomenclature{ZrH}{Zirconium hydride}
\nomenclature{Be}{Beryllium}

\nomenclature{HFP}{hot full power}
\nomenclature{HZP}{hot zero power}
\nomenclature{CPU}{Central processing unit}

\printnomenclature


\newcolumntype{L}[1]{>{\raggedright\arraybackslash}p{#1}}
\renewcommand{\hl}[1]{#1}

\section{Introduction}
\label{sec:intro}
Microreactors ($\mu$R) are a type of nuclear reactor characterized by their low power outputs (below 20 MWth \cite{naranjo2024assessment}), compact designs, transportability (e.g., by truck or train), that they can afford onsite installation, factory assembly, and inherent self-regulation (i.e., the system requires no human intervention or backup power to shut down safely) \cite{choi2024advancements,park2025bottom,naranjo2024assessment}. $\mu$R refueling is envisioned to be conducted at a central facility, eliminating the need to handle and store used fuel at each individual location. These characteristics open up opportunities in non-traditional market segments, including integration within complex architectures serving as heat sources for hydrogen electrolyzers \cite{seurin2022h2goldenretrievermethodologytoolevidencebased,SEURIN2026111865}, district heating and cooling, synthetic fuel production, and desalination \cite{buongiorno2021can}, as well as applications in space reactors, deep-sea environments, remote polar regions, mines, and military bases \cite{bryan2023remote,wang2025data}. $\mu$Rs are expected to produce on-demand, dispatchable, and reliable energy, and thus are gaining traction as a key component of the U.S.~Government's ambitious strategy to ensure American energy dominance and economic prosperity \cite{goff2025key}. More recently, they have attracted the interest of hyperscalers and data center stakeholders, both of whom require uninterrupted service and excellent voltage and frequency stability \cite{aljbour2024powering,seurin2025longterm} so as to meet the recent power demand surge caused by the exponential rise in artificial intelligence (AI) usage \cite{anderson2024microgrid,seurin2025longterm}, as well as to ensure AI leadership in support of national security missions, thereby bolstering U.S.~economic competitiveness.

Heat-pipe microreactors (HPMRs) are particularly well-suited for many of these applications, thanks to their inherent safety features. Additionally, owing to their often low power density, they can operate for more than 5--10 years, making them the preferred choice for deep space exploration and remote, unattended operation--- applications that could result in improving their economic value propositions \cite{wang2025data}. HPMRs utilize heat pipes (HPs) to transfer heat from the reactor to the power conversion system, and Westinghouse is currently developing a model at scale via its eVinci$^{TM}$ $\mu$R program \cite{price2024multiphysics}. A type of isothermal exchanger, HPs are ubiquitous in contemporary electronics. The advantage of HPs is that they do not require any circulation pumps or auxiliary systems, thus theoretically enabling greater portability and cost-efficiency \cite{yan2020technology}.


Although diseconomies of scale may cause the $\mu$R to experience higher costs per unit of energy output as compared to large nuclear plants, several compensatory benefits can be anticipated, including standardization, simplification, passive safety, reduced radionuclide inventories, factory-based fabrication, rapid installation, and lower financing costs \cite{abou2021economics}. While $\mu$R were originally proposed for large markets \cite{buongiorno2021can}, they have yet to fully capitalize on this strategy, as they still do not yet offer a reasonable cost of energy. 

The economic competitiveness of $\mu$R varies greatly depending on the market and customers. Currently, they are more targeted at markets in which they could replace expensive small- to medium-sized fossil fuel turbines and diesel generators \cite{naranjo2024assessment}, and in which carbon abatement regulations (e.g., carbon taxes, cap-and-trade programs, or emission limits) \cite{buongiorno2021can} and policy incentives including clean energy production tax credits \cite{abdusammi2025evaluation} could serve to enhance their competitiveness \cite{al2025open,buongiorno2021can}. The intermittency of renewable energy sources often makes them poorly suited to such markets, and many applications cannot tolerate service disruptions or the additional costs of maintaining storage capacity \cite{buongiorno2021can}. Nor is it easy to meet the large land requirements involved \cite{naranjo2024assessment}. Recent core design optimization approaches for large water reactors \cite{seurin2024assessment,seurin2025surpassing} and small modular reactors \cite{halimi2024scale,halimi2025fuel,erdem5868125maeo} have focused on loading pattern optimization with legacy lattice geometries (e.g., Westinghouse \cite{seurin2024assessment,halimi2024scale}), as well as design of assemblies featuring varying enrichments and burnable poison patterns. This has limited the number of design configurations that can offer an integer input space. On the other hand, the $\mu$R design space is significantly larger---with a continuous search space---thus making hand-design prohibitive. As nuclear vendors are still identifying what markets are appropriate for these reactors, optimization within this space must be explored.

Rather, the focal point for economic analysis has been on fixing or limiting the geometric design parameters and instead focusing on cost scoping analysis \cite{buongiorno2021can,abdusammi2025evaluation}, market analysis \cite{buongiorno2021can,park2025bottom} enhanced fuel loading options to improve fuel cycle costs \cite{al2026assessing,park2025bottom}, scoping analyses for material design choices \cite{shirvan2023uo2,hanna2025bottom}, increased automation \cite{naranjo2024assessment,bryan2023remote}, power uprates \cite{park2025bottom,hanna2025bottom}, and multi-unit options with centralized refueling facilities (i.e., co-siting and equipment sharing to accelerate (1) learning in the area of nuclear equipment and manufacturing and (2)reduced licensing requirements) \cite{park2025bottom,bryan2023remote}. On the other hand, several studies have focused on HPMR geometric design optimization by using multiphysics coupling to simultaneously optimize power density and fuel enrichment \cite{wang2025data}, or the optimization of thermal stress distribution so as to ensure structural safety \cite{zhang2025optimizing}. Both these approaches leverage the legacy multiobjective genetic algorithm known as the Non-dominated Sorting Genetic Algorithm (NSGA-II) \cite{deb2002fast}. The difficulty in performing comprehensive optimization hinges on the complex neutronic, thermal, and mechanical interactions inherent to HPMR designs \cite{zhang2025optimizing}, evaluations of which necessitate expensive codes (e.g., OpenMC \cite{romano2015openmc}) and, oftentimes, surrogate models. Moreover, they assume that the design are economically competitive but many analysis have demonstrated that HPMRs are one of the least ones \cite{shirvan2023uo2}.

To the authors' knowledge, no previous study ever conducted a unifying optimization effort that measures the impact of geometric design decisions on HPMR economic assessments. Democratized access to techno-economic tools such as Microreactor Optimization Using Simulation and Economics (MOUSE) \cite{hanna2025bottom}, open-source packages for Monte Carlo simulation (e.g., OpenMC \cite{romano2015openmc}), surrogate modeling (e.g. scikit-learn \cite{scikit-learn}), and optimization via the Neuroevolution Optimization with Reinforcement Learning package \cite{radaideh2023neorl} can aid us in achieving this. More specifically, recent advances in AI---particularly as regards reinforcement learning (RL) \cite{seurin2024multiobjective}---have demonstrated the potential to rapidly and efficiently characterize safety and economic performance trade-offs in the design of large-scale light-water reactors.

The primary objective of this work is to introduce a method for optimizing the economy of $\mu$Rs in the presence of computationally intensive physics codes by utilizing a combination of surrogate modeling and automated optimization tools. The secondary objective is to understand what geometric design decisions may lead to more cost-efficient reactors. We will present two use cases in which the cost of the axial reflector (i.e., beryllium (Be)) varies between its nominal cost (i.e., 45,000 \$/kg in \$2024 USD) and the cost of graphite (i.e., 80 \$/kg in \$2022 USD), thereby greatly influencing the topology of the optimal designs. A companion paper will address the inherent correlation between safety and economics by employing a multi-objective RL-based algorithm (i.e., the Pareto Envelope Augmented with Reinforcement Learning \cite{seurin2024multiobjective}). The contribution of this work can be summarized as follows:

\begin{enumerate}
\item First-of-a-kind HPMR optimization with techno-economic analysis considerations.
\item Identification of the major HPMR cost drivers, and guidance on research directions for achieving a more economically viable design, as supported by recent advances in RL.
\end{enumerate}

The remainder of the paper is organized as follows. In Section \ref{sec:methodology}, we present the nominal design on which we will base our optimization (see Subsection \ref{sec:nomdes}), the physics modeling and techno-economic modeling applied to our candidate evaluation (see Subsection \ref{sec:physicsandtechnoecon}), the design parameterization and bounds (see Subsection \ref{sec:designparm}), and the optimization methodology, including the formulation of the objective function with RL 
(see Subsection \ref{sec:optimiztionmethodoogy}). The results are then given in Section \ref{sec:res}, along with a description of the data generated for the surrogate model (see Subsection \ref{sec:samplesdescription}), its performance on these data (see Subsection \ref{sec:surrogatemodel}), the integration of RL 
in analyzing the best designs found (see Subsection \ref{sec:applicationof}), and the deep dive that was conducted into the levelized cost of electricity (LCOE) for a First-of-a-kind (FOAK) design (see Subsection \ref{sec:deepdive}). Lastly, concluding remarks and areas of future research are provided in Section \ref{sec:conc}.

\section{Methodology}
\label{sec:methodology}

\subsection{Nominal Reactor Design}
\label{sec:nomdes}

\begin{figure}[htb!]
    \centering
    \includegraphics[width=.9\textwidth]{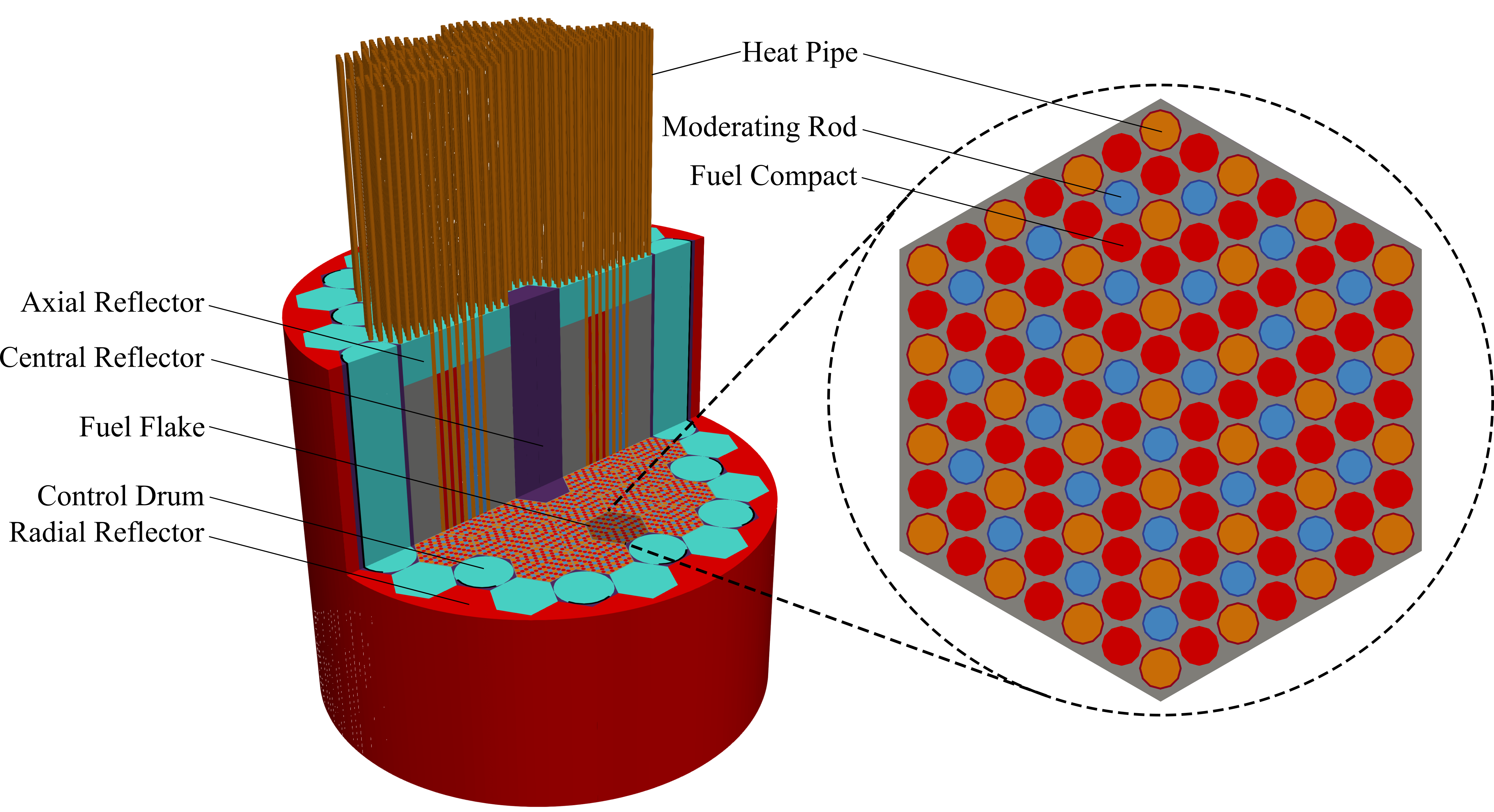}
    \caption{Nominal reactor design, with labeled components (subject to change during the optimization process).}
    \label{fig:rctr_diag}
\end{figure}

The HPMR design used in this study was based on a modeling experiment for demonstrating advanced modeling and simulation capabilities \citep{stauff2022multiphysics}.
As this design was originally made without real-world implementation in mind, it is a particularly useful candidate for the current methodology as it should theoretically afford room for design improvements.
Both Figure \ref{fig:rctr_diag} and Table \ref{tab:nominal_design_characteristics} list the primary characteristics of the nominal design. Current HPMR concepts have not prioritized economic considerations, but rather technological readiness as well as testing of the Nuclear Energy Advanced Modeling and Simulation program's capabilities in order to develop faithful, high-fidelity advanced reactor models \cite{osti_1891623}. As discussed in Subsection \ref{sec:designparm}, some aspects of this design are subject to change in the process of seeking a more optimal design.
Subsection \ref{sec:objective_fxn_formulation} mathematically defines the term ``optimal design'' as it is used in the context of the present study.

\begin{longtable}[c]{L{5cm} L{8cm}}
\caption{Major design characteristics of the nominal core design.}
\label{tab:nominal_design_characteristics}\\
\hline
Design Specification       & Information            \\ \hline
\endfirsthead
\multicolumn{2}{c}%
{{\bfseries \tablename\ \thetable{} -- continued from previous page}} \\
\hline
Design Specification       & Information            \\ \hline
\endhead
\hline \multicolumn{2}{|r|}{{Continued on next page}} \\ \hline
\endfoot
\hline
\endlastfoot
Core Characteristics       & \begin{tabular}[t]{@{}l@{}}
Power -- 2 MWth \\
Total height -- 200 cm\\ Active fuel length -- 260 cm\\ Reactivity control -- 12 control drums\end{tabular} \\
Fuel Flake Loading         & \begin{tabular}[t]{@{}l@{}}Pin pitch -- 3.2 cm (flat-to-flat)\\ No.~fuel compacts -- 63 per flake\\ No.~HPs -- 37 per flake\\ No.~moderator rods -- 27 per flake\end{tabular}                          \\
Fuel Flake Characteristics & \begin{tabular}[t]{@{}l@{}}Total no.~-- 30\\ Flat-to-flat width -- 26.752 cm\\ Material -- graphite\end{tabular}                                                                                            \\
Fuel Compacts              & \begin{tabular}[t]{@{}l@{}}Description -- TRISO particles in graphite matrix\\ Fissile material -- 70 at\% UO$_2$; 30 at\% UC\\ Enrichment -- 19.7\%\\ Packing fraction -- 40\%\\ Radius -- 1 cm\end{tabular} \\
HPs                 & \begin{tabular}[t]{@{}l@{}}Envelope outer radius -- 1.05 cm\\ Envelope thickness -- 0.8 mm\\ Envelope material -- 316 stainless steel\\ Wick thickness -- 1 mm\\ Working fluid -- potassium\end{tabular}    \\
Moderator Rods             & \begin{tabular}[t]{@{}l@{}}Cladding outer radius -- 0.92 cm\\ Cladding thickness -- 0.095 cm\\ Moderator outer radius -- 0.825 cm (including gap)\\ Moderator material -- yttrium hydride\end{tabular}         \\
Radial Reflector           & \begin{tabular}[t]{@{}l@{}}Flat-to-flat width -- 260 cm\\ Material -- graphite\end{tabular}                                                                                                                    \\
Axial Reflector            & \begin{tabular}[t]{@{}l@{}}Thickness (top and bottom) -- 20 cm\\ Material -- Be\end{tabular}                                                                                                             \\
Control Drums              & \begin{tabular}[t]{@{}l@{}}Coating material -- enriched boron w/ 95 at\% B-10\\ Coating angle -- 90\textdegree\\ Coating thickness -- 1 cm\\ Outer diameter -- 26.5 cm\end{tabular}                              \\ \hline
\end{longtable}

\subsection{Physics and Techno-economic Modeling}
\label{sec:physicsandtechnoecon}
The two major modeling components in this work are physics and techno-economic modeling.
Physics modeling is necessary to estimate quantities of interest (QoIs) such as the reactor lifetime, peaking factor, and shutdown margin (SDM); techno-economic modeling is necessary for $\mu$R financial cost/benefit analyses.

\subsubsection{Physics Modeling}
In this work, the Monte-Carlo-based OpenMC \cite{romano2015openmc} continuous-energy neutron transport code was used to model the candidate designs.
The ENDF/B-VIII.0 cross-section libraries were also employed \cite{brown2018endf}.
The Chebyshev rational approximation method-based \cite{maria2016higher} depletion capabilities afforded by OpenMC enable core lifetime estimations, and the stochastic tri-structural isotropic (TRISO) particle dispersion routines allow for modeling the random distributions of TRISO particles within the fuel compacts.

A few minor modeling details can now be discussed.
First, the fuel composition is lumped into 5 depletion zones.
Within each zone, the composition of the fuel within the TRISO particle kernel is homogenized with all the other fuel materials in that same zone.
The 15 depletion zones are created with three radial divisions and five axial divisions imposed on the core model.
The radial divisions differentiate the fuel in each of the three rings of hexagonal flakes that comprise the core.
This depletion zoning will yield better lifetime estimate accuracy than approaches that only track a single fuel composition across the entirety of the core.
Furthermore, only a single particle dispersion routine is run for a volume comprising one-fifth of the axial length of a single compact. 
The generated dispersion is then vertically stacked five times to form the dispersion pattern for all compacts in the core.
Aside from some minor artifacts in the axial flux shape that may impact highly-detailed design analyses (which are beyond the scope of this article), this approach is not expected to significantly impact the results.
In general, non-homogenization-based assumptions about TRISO particle distributions tend to have a minor impact on the evolution of k$_\infty$ throughout the core lifetime \cite{impson2025influence}.

A few minor tallying/postprocessing details should also be discussed.
First, given that peaking factors are calculated using a Monte Carlo code, the neutron flux cannot be estimated on a point-wise basis.
Instead, it must be estimated as an average within a particular volumetric tally region.
For the results shown in this work, each fuel compact is divided into 20 axial tally regions.
For pin-wise peaking factors, this has no effect, but for axial peaking it leads to underestimation.
However, this underestimation is expected to be relatively consistent across all the candidate designs and will therefore not impact the demonstration covered in this article.
Next, the core lifetime is calculated by estimating the time that k$_{\text{eff}}=1$.
Many candidate designs in the search domain are non-starters, meaning that k$_{\text{eff}} < 1$ even for a fresh core.
To provide the optimization algorithm with feedback on exactly how ineffective a particular non-starter candidate design is, the core lifetime is assigned a negative value based on a backward linear extrapolation so as to find the negative time value where k$_{\text{eff}}=1$.
This backward extrapolation uses the first two burnup steps---discounting the very brief initial step taken to reach quasi-equilibrium in short-lived isotope concentrations.

\subsubsection{Techno-economic Modeling}

The economic competitiveness of each design can be evaluated using bottom-cost estimate tools \cite{park2025bottom,al2025open} in tandem with the code-of-account (COA) methodology, which calculates the total cost of each individual component, including the fuel cost, operation and maintenance (O\&M) cost (e.g., labor, spare parts, taxes, and fees), and indirect/direct capital costs. Open-source COA tools offer opportunities for collaboration, transparency, and consistency when performing such analyses. One such COA tool is MOUSE, which is currently being developed at Idaho National Laboratory \cite{hanna2025bottom}, in an open-source package hosted in the  \href{https://github.com/IdahoLabResearch/MOUSE}{IdahoLabResearch} GitHub repository, and which is utilized here for the sake of reproducibility. The MOUSE methodology is based on prior work started in \cite{al2025open,abou2021economics}. Most of the baseline cost assumptions are reported in the associated references. The lack of construction cost data, stemming from the absence of $\mu$R builds and guidelines on conducting techno-economic assessments of these assets, introduces inherent uncertainties in their actual values. The cost that informs the optimization is obtained by propagating all these uncertainties and then taking the mean.

In \cite{hanna2025bottom}, reactor physics, shielding, and thermal design principles were analyzed for an HPMR nominal design via an original techno-economic assessment. We will use the same assumptions for our design described in Subsection \ref{sec:nomdes}. For the purposes of this study, we recall several factors that will be affected by our design choices, as presented in Subsection \ref{sec:designparm}. First, we assume that the $\mu$R will be shipped, fueled, operate continuously (except for several emergency shutdowns, which will be captured in the capacity factor), and then be shipped back to a central facility for refueling and servicing (e.g., repair, maintenance, or replacement). For this model, the LCOE was obtained by summing the discounted fuel, O\&M, and capital costs over the lifetime of the $\mu$R, then dividing it by the discounted energy produced over its lifetime:
\begin{equation}
    LCOE = \frac{\sum_{t=0}\frac{F_t+O_t+TCI_t}{(1+r)^t}}{\sum_{t=0}\frac{E_t}{(1+r)^t}},
\end{equation}
where $F_t$ is the cost of fuel in year t; $O_t$ is the O\&M cost in year t; $TCI_t$ is the total capital invested (TCI), defined as the sum of the overnight capital cost (OCC) and the financial cost (e.g., interest, escalation, and contingencies), in year t; $E_t$ is the generated electrical energy in year t, r is the 
discount rate, and n is the expected plant lifetime in years. Here, r is assumed to be 6\%, while n is assumed to be 60 years. Note that $TCI_t$ occurs before and during construction, and thus is only included in the first term ($t=0$), whereas the other two are not accounted that time and $E_0 = 0$.

\begin{enumerate}
    \item $F_t$ encapsulates the costs of the natural uranium, conversion, enrichment, and fuel fabrication, all summed up over a refueling interval. The costs are assumed to be incurred at the start of the irradiation cycle. The cost of the fuel is assumed to be 10,000 \$/kgU  \cite{hanna2025bottom,abou2021economics} (in \$2009 USD, as per \cite{williams2021advanced}), no matter the fuel content (whether $UO_2$, $UCO$, $UN$, or $UC$). However, it would be expected that for metallic fuel the cost would slightly increase. For the present work, we used $UCO$. The mean costs for the other rubrics are reported in Table \ref{tab:costforfuelcycle}.
    \begin{table}[htp!]
        \centering
        \caption{The fabrication cost is given in Table \ref{tab:costperkganddensity} in \$2009 USD. The term 1.15 is a penalty provisioned for enrichment levels above 10\%.}
        \begin{tabular}{lll}
        \hline
    &  Mean cost (\$2022 USD)  & Unit  \\
        \hline
          Natural Uranium   & 184 & \$/kgU \\
           Conversion  & 15.1 & \$/kgU\\
           Enrichment & 184.2 × 1.15 & \$/SWU (Separation Work Unit)\\
          \hline
        \end{tabular}        \label{tab:costforfuelcycle}
    \end{table}
    Despite recent discussions regarding the benefit of costing the spent fuel disposal based on fuel volume rather than on the per unit of megawatt produced \cite{kim2022nuclear,halimi2024scale}, the cost of spent fuel disposal will be assumed to be 1 \$/MWh (as a best practice \cite{seurin2024assessment,shirvan2023uo2,buongiorno2021can}), meaning based on the energy produced rather than on the spent fuel volume \cite{halimi2024scale}. Note that this is similar to the cost of decommissioning assumed to be 1,100 \$/KWe. The tailing is assumed to be 0.25\% and the natural uranium 0.71\%.
    \item $O_t$ encompasses the personnel labor costs incurred for security, operation, inspection, and servicing, as well as the cost of replacing or maintaining operational spare parts. 
    In the ``autonomous'' option that has been chosen going forward, the reactor is monitored remotely and operators are required onsite only in the event of emergency shutdown or startup \cite{hanna2025bottom}. It is is assumed that the startup following an emergency shutdown requires two operators, each with a 10-hour-per-day schedule, for a full-time equivalent (FTE) of approximately 0.08. The resulting cost is thus $178,500 \times 0.08 \sim \$14,000$ per operator. Monitoring is conducted at a central facility by a single operator who is responsible for 10 reactors. Under the assumption that 5 FTEs (1 operator 24/7) are spent on monitoring these reactors, the cost amounts to $178,500\times 5 \times \frac{1}{10}\sim 90,000\$$. Security must, however, still be onsite 24/7, which is equivalent to 5 FTEs, for a total of 178,500\$/FTE (in \$2024 USD) \cite{al2025open}.  
    
    The maintenance cost depends on the equipment utilized. The main components, including the vessels, reflector, moderator, and control drums, are replaced approximately every 10 years (based on the Advanced Test Reactor timeline \cite{hanna2025bottom}). Note that for the cost calculation, the drum height is equal to the fuel height $x_{fh}$. For the other components, an annual cost of 1.5\% relative to the direct cost is assumed. The original weight of each component therefore impacts the maintenance costs andlarger drums or a wider moderator radius (which replaces lower-density graphite) incur a penalty on the maintenance cost. The maintenance is performed during refueling and thus the capacity factor ($C_f$) is upper bounded by its value when the fuel lifetime is 10 years. $C_f$ depends on the fuel lifetime, refueling period, and number and duration of emergency shutdowns per year. Similarly, the capital plant expenditures represent the expected cost incurred by upgrades to maintain or improve plant capacity, meet future regulatory requirements, or extend plant lifetimes. They are a percentage of the total capital direct cost \cite{al2025open}, which is itself dependent on the contribution of each of these components in the reactor.
    
The cost assumptions for each rubric are listed in Table \ref{tab:capacityfactorassumptions}.
    \begin{table}[htp!]
        \centering
        \caption{Baseline assumptions used in evaluating the capacity factor and cost of refueling operations.}
        \begin{tabular}{lll}
        \hline
           Rubrics  & Values & Units \\
           \hline
          Number of Operators 
(required for an emergency or refueling) &	2 & -\\
Levelization Period &	60 & years \\
Refueling Period &	7 & days\\
Number of Emergency Shutdowns per Year &	0.2 & -\\
Startup Duration After Refueling &	2 & days\\
Startup Duration After Emergency Shutdown & 14 & days\\ 
Number of Reactors Monitored per Operator &	10 & -\\
Number of Security Staff Per Shift	& 1 & - \\
\hline
        \end{tabular}
        
\label{tab:capacityfactorassumptions}
    \end{table}
 Other costs include regulatory fees (Nuclear Regulatory Commission operating and inspections fees), property taxes, and insurance premiums---none of which are affected by our design changes.
 
 \item  The direct cost component in the OCC is estimated using the cost-to-capacity method, which scales the cost of a system when adding capacity (mass or power) to it:

\[I_o = I_{fixed} + I_{ref}\left(\frac{X_0}{X_{ref}}\right)^{n_{\text{scale}}},\]

where $I_{fixed}$ is the fixed cost, $I_{ref}$ is the reference cost, $X_0$ is the scaling variable, $X_{ref}$ is the reference value of the scaling variable, and $n_{\text{scale}}$ is a scaling factor obtained from historical data and depends on the system. The scaling exponent value is generally 1 when the scaling variable related to the mass of any component, and ranges between 0.6 and 0.8 when the scaling variable is thermal or electric power. Although this method was used to scale costs for most components, certain costs were scaled using specific equations taken from the literature. These specific scaling equations were applied to the costs of the pumps, compressors, heat exchangers, and fuel enrichment.

The control drums are an important aspect of the design parameters and significantly contribute to the cost. The cost can be decomposed by the cost per drum and the loading of Be and boron carbide ($B_4C$): \[ m_{Be} C_{Be} + m_{B_4C}C_{B_4C} + (I_{installation} + I_{fabrication}) N_{CD},\] where $C_{Be} = 45,000$ \$/kg in \$2024 USD, and $C_{B_4C} = 14,268$ \$/kg  in \$2023 USD for the natural $B_4C$ (around 20\% $B_{10}$) and 10,064 \$/kg for the enriched $B_4C$. In practice, the cost of $B_4C$ should vary depending on its enrichment. But to date, no rigorous approach exists for evaluating its exact cost. Since we are working with a high enrichment level, we will use the latter of the two costs. Moreover, the installation and fabrication costs, which amount to $I_{installation} = 80,665$ in \$2024 USD and $I_{fabrication} = 323,650$ in \$2024 USD, respectively, and which originate from the Marvel data, will also be affected by the size of the control drums.
    
Furthermore, due to the importance of HPs, we recall that the cost is assumed to be 10,000 \$/HP \cite{abou2021economics} in \$2017 USD. This is valid here, as the number and size of the HPs will not change in our design. Additionally, the TRISO fuel, graphite moderator, and the yttrium hydride (YHx) solid moderator booster have different densities and costs per kg. Since we will change the radius of the compact and moderator, both the OCC and the maintenance portion of the $O_t$ will be impacted (see Table \ref{tab:costperkganddensity}).
    
    \begin{table}[htp!]
        \centering
        \caption{Density and cost per kg for the HPMR core materials. The cost of YHx is assumed to be similar to that of zirconium hydride (ZrH), though in actuality it is likely higher. However, its impact on the LCOE is negligible \cite{hanna2025bottom}.}
        \begin{tabular}{lll}
        \hline
          Materials & Cost (\$/kg) & Density (g/$cm^3$)\\
          \hline
           TRISO & 10,000 (\$2009 USD) & 3.2--3.5\\
           YHx & 1,520 (\$2017 USD) & 4.3--4.6 \\
           graphite & 80 (\$2022 USD) & 1.9--2.3\\
           \hline
        \end{tabular}
        
\label{tab:costperkganddensity}
    \end{table}
The indirect costs are difficult to evaluate and are often assumed to correlate to the direct costs. Interested readers can refer to Section 3.9 in \cite{al2025open} for details or the recent work in \cite{candido2025benchmarking} for a more granular itemization of the indirect costs.
Lastly, the TCI is obtained by summing the OCC and the capitalized financing cost (account 60), with an assumed debt-to-equity ratio of 0.5.
\end{enumerate}

Mass production is the most important pathway via which $\mu$R vendors can achieve economic competitiveness, as enabled by two mechanisms.
The first mechanism is mass manufacturing, which reduces costs thanks to bulk purchasing. Cost multipliers are adopted to adjust the cost of equipment/tasks in the context of mass production via a factory setup. These multipliers depend on the type of equipment/task at hand. The second mechanism is onsite learning (continuous improvement stemming from the learning and innovation accrued between FOAK and Nth of a kind (NOAK)). The NOAK cost of building $N$ $\mu$Rs units is estimated after 20 units have already been built, and assumes that no further learning occurs after the 100th unit \cite{hanna2025bottom}:
\begin{equation}
    NOAK = FOAK \times (1 - lr)^{\log_2 N},
\end{equation}
where $lr$ is the learning rate, which depends not only on the type of equipment but also on the specific account. The learning rate in the context of factory activities is assumed to exceed that of onsite activities. Some accounts do not experience learning, as they are already mass produced or would not benefit from mass production. Examples of these are land costs, permits, and plant studies. 
Since FOAK is what is considered during the optimization, uses of ``LCOE'' throughout the rest of this article will refer to the FOAK LCOE unless ``NOAK'' is otherwise specified. 

Lastly, data provided in any measure other than \$2024 USD are escalated to \$2024 USD.

\subsection{Design Parameterization and Bounds}
\label{sec:designparm}

Through the optimization process, a number of modifications were made to the nominal core design, introduced in Subsection \ref{sec:nomdes}, in order to improve certain reactor performance objectives.
Conceptually speaking, these modifications were made by parameterizing the model---or equivalently, by expressing a particular design in terms of a vector of design parameters.
All discrepancies between the new candidate design and the nominal one should be inferrable from this vector alone.
Bounds can also be introduced to these parameters to ensure that the candidate designs meet the practical system constraints, and to limit the range of designs being considered.

Table \ref{tab:inparms} introduces seven parameters, along with their nominal design values, bounds, and mathematical symbols.
To ensure geometric viability, the upper bounds of the compact and moderator radii are limited the pin pitch.
All other bounds are then selected with the intention of encompassing the parameters associated with the optimal design.
Further discussion is required before a complete model can be formed around these seven parameters.
First, the flat-to-flat width of the flake is scaled to maintain the additional width that sits outside the geometric pin lattice in the nominal design.
Specifically, the flake width for a candidate design can be calculated as $\frac{13\sqrt{3}}{2}x_{pp} + 0.858$.
Second, the total volume of the core is unchanged, thus satisfying concerns regarding the transportability of the candidate designs \cite{buongiorno2021can,shirvan2023uo2}.
The axial and radial reflectors will shrink and grow so as to maintain the outer boundary of the core.
The control drum radius is calculated based on the size of the flake unit cell in order to maintain the distance between the control drum outer surface and the edge of the flake unit cell present in the nominal model.
With this, the control drum outer radius is calculated as $\frac{1}{2}\left(\text{flake width} - 0.252 \text{ cm}\right)$.
The control drum coating thickness remains at 1 cm, regardless of drum size.
Finally, the fuel packing fraction is kept constant from one candidate design to the next.

\begin{table}[htb!]
\centering
\caption{Parameter descriptions for changing aspects of the candidate designs.}
\label{tab:inparms}
\begin{tabular}{llll}
\hline
Symbol    & Description                           & Nominal Value & Bounds                                                                                                     \\ \hline
$x_{ca}$  & Control drum coating angle            & 90$^\circ$     & [35, 180]                                                                                                  \\
$x_{B10}$ & Control drum absorber B-10 enrichment & 95\%         & [20\%, 95\%]                                                                                               \\
$x_{fh}$  & Active fuel height                    & 160 cm        & [130 cm, 190 cm]                                                                                           \\
$x_{pp}$  & Pin pitch                             & 2.3 cm        & [1.94 cm, 2.78 cm]                                                                                         \\
$x_{e}$   & U-235 fuel enrichment                 & 19.7\%        & [17\%, 19.9 \%]                                                                                             \\
$x_{cr}$  & Fuel compact radius                   & 1 cm          & $\left[\frac{1}{4}x_{pp}, \frac{1}{2}x_{pp}\right]$                                                                 \\
$x_{mr}$  & Moderator radius                      & 0.825 cm      & $\left[\frac{1}{5}\left(x_{pp}-2(0.095 \text{ cm})\right), \frac{1}{2}\left(x_{pp}-2(0.095 \text{ cm})\right)\right]$ \\ \hline
\end{tabular}
\end{table}

HPMR optimization is inherently multiphysics, involving a complex web of neutronic, thermal, and mechanical feedback interactions \cite{zhang2025optimizing}. We cannot account for all these interactions solely with a Monte Carlo code; however, we can consider proxies that would be able to accommodate them (e.g., by using peaking factors)--important QoIs related to cost and neutronics were extracted from OpenMC, including the fuel lifetime, SDM, and temperature coefficients (TC)s. The final set of QoIs we considered are as follows:
\begin{enumerate}
    \item 
\textbf{Lifetime}: Operational time or fuel lifetime. 

\item \textbf{SDM}: Expressed in pcm, this is a measure of the margin available for shutting down the reactor from operating conditions. In addition to the available control rods (or drums in our case), nuclear reactors must be designed to have negative TCs. The reactivity change from hot full power (HFP) to hot zero power (HZP) is large, resulting in a penalty to the margin. Though there are multiple ways to evaluate this, we will adopt the methodology from \cite{halimi2024scale}:
\begin{equation}
    SDM = \Delta k_1 + 0.9 \times (\Delta k_2 - \Delta k_3),
\end{equation}
where $\Delta k_1$ is the reactivity difference between HZP to HFP, $\Delta k_2$ is the HFP to ``all rods in,'' and $\Delta k_3$ is the HFP to ``most effective rods in.'' $\Delta k_1$ will be evaluated by setting all temperatures across the core equal to the working fluid (e.g., sodium or potassium) operating temperature in the HP (here 800 K). For pressurized-water reactors, control rods (CRD)s operate in banks to ensure a symmetric attenuation of the neutron flux density and, thus, the power profile in the core. In the HPMR case, due to the size and location of the control drums, they are assumed to operate with independent rod drives. As a result, $\Delta k_3$ can be computed by rotating only a single drum out of the core. This drum can be arbitrary due to the axis-symmetry of our reactor core and the fact that it is one-through (i.e., the entirety of the core is replaced upon discharge). 

\item \textbf{Temperature coefficients}: Nuclear reactors must have negative TCs to ensure that a power increase will stabilize the chain reaction. In an HPMR, the competing effects include Doppler broadening (i.e., the rise in neutron absorption cross-sections) and the temperature feedback of the HP walls (i.e., reduction in capture cross-sections). In some designs, the Doppler feedback dominates over the other TCs \cite{wang2025data,choi2024advancements}, but caution must be exercised in the presence of solid moderators such as YHx.

For the many $\mu$R that will operate at temperatures above 400$\degree$C, use of water as a moderator is restricted due to the extremely high pressure required. As an alternative to improve compactness, solid moderators such as ZrH and YHx are utilized. Recently, YHx has been considered a more viable option, thanks to its higher hydrogen stability under temperature and irradiation, as well as to its attainable hydrogen density \cite{hu2020fabrication,hanna2025bottom}. However, designs based around YHx moderators are known to carry the risk of exhibiting positive TCs due to reduction in Y capture cross-sections \cite{Ade02122022}. Under certain conditions, this can lead to reactor instability. The isothermal TC (ITC) can be evaluated as a worst-case scenario. Indeed, since the Doppler effect will activate first, evaluating the $k_{eff}$ that results from an isothermal change in reactor temperature will provide an upper bound on the TC. The change in reactivity $\Delta \rho_{TC}$ from $T_1$ to $T_2$ can then be evaluated by:
\begin{equation}
    \Delta \rho_{TC} = \frac{k_{T_1} - k_{T_2}}{k_{T_1}k_{T_2}},
\end{equation}
where TC is the temperature coefficient. The ITC is then $ITC = \frac{\Delta \rho_{TC}}{T_1 - T_2}$. It will be evaluated at 550--850 K and at 850--1150 K, as its value can change depending on the temperature \cite{Ade02122022}. During depletion, the spectrum will harden and the coefficient could become less negative. The ITC must therefore be evaluated at multiple depletion steps, substantially lengthening the collection of data. As a result, the ITC will be evaluated only a posteriori for the final designs (see Subsection \ref{sec:applicationof}).
\item 
\textbf{Peaking factors:} 
Being the central components of the reactor, HPs are subject to various failure modes---including boiling, entrainment, sonic, viscous, and capillary limits---that could compromise their heat removal capabilities. When an HP fails, the reactor does not necessarily shut down; instead, the heat load on the neighboring HPs increases, potentially leading to further failures that could propagate throughout the reactor. This phenomenon is known as ``cascading HP failure.'' If the heat sustained by the remaining HPs becomes excessive, the designer must make the necessary adjustments. Among the additional design considerations that arise from prolonged increases in local heat load and power are thermal expansion, thermal creep, and corrosion effects all of which may jeopardize HP integrity. As a proxy for these power- and temperature-induced effects, we will consider two peaking factors in this study: $F_q$, the maximum pin peaking factor (to obtain the peak heat flux $q^{''}_{max}$), and $F_{\Delta h}$, the maximum rod-integrated peaking factor \cite{seurin2024assessment}. 

$F_q$ is connected to the linear heat generation rate by \cite{al2023design}:
\begin{equation}
    F_q = \frac{q'_{max}}{q'_{avg}}.
\end{equation}
The average heat flux from one compact is obtained as $q^{''}_{avg} = \frac{Q}{N_{flakes} \times N_{compact} \times L \times 2 \pi r} = \frac{q^{'}_{avg}}{2\pi r}$, where $N_{flakes}$ is the number of flakes in the core, $N_{compact}$ is the number of compacts per flake, $r$ is the radius of the compact, and $L$ is the active length of the fuel. In this work, we utilize TRISO particles arranged in a compact configuration, allowing the heat flux to be calculated as an average flux emanating from each compact at 20 different axial locations. The peak heat flux is then obtained as follows: $q^{''}_{max} = F_q\times q^{''}_{avg}$. Then we also have:
\begin{equation}
    F_{\Delta h} = \frac{\max_{x,y} \int_0^L q_{x,y}^{'}(z)dz}{\frac{1}{N_{compact}}\sum_{x,y}\int_0^L q_{x,y}^{'}(z)dz},
\end{equation}
where $x$ and $y$ define the radial location of each compact. 
The product of $q^{''}_{max}$ and $F_{\Delta h}$ can be interpreted as a proxy for the maximum temperatures of the fuel, monolith, reflector, solid moderator, and HP, as well as for the maximum energy transferred to the HP.
\end{enumerate}

The QoIs associated with the nominal design are presented in Table \ref{tab:nominaldesignobjectives}.
\begin{table}[htp!]
    \centering
    \caption{QoIs associated with the nominal design described in Section \ref{sec:nomdes}. The LCOE for the nominal design is given for both the cost of the nominal Be and that of graphite, with the two values being separated by a slash.}
    \begin{tabular}{lll}
    \hline
    QoIs & Values & units \\
    \hline
      Lifetime   & 6.99 & years\\
                
    SDM & - 6757.23 &  pcm\\
                
$Fq$ &1.787 & - \\
$F_{\Delta h}$&1.469 & - \\
$q^{''}_{max}$ & 0.0188 & MW$/m^2$ \\
ITC & -2.404$^{\star}$ & pcm/K \\
LCOE (FOAK) Estimated Cost (\$2024 USD)&10,307/5,079& \$/MWh\\
LCOE (NOAK) Estimated Cost (\$2024 USD)&1,596/1,442& \$/MWh\\
\hline
    \end{tabular}
\label{tab:nominaldesignobjectives}
\begin{tablenotes}
    \item$^{\star}$ Worst-case scenario for the temperature range 850--1150 K.
\end{tablenotes}
\end{table}


\subsection{Optimization Methodology}
\label{sec:optimiztionmethodoogy}
\subsubsection{RL for Optimization and Objective Function Formulation}
\label{sec:objective_fxn_formulation}

RL was introduced to overcome the limitations of classical heuristic-based approaches in optimizing complex engineering problems-- approaches that typically rely on navigating the search space by employing stochastic rules of thumb \cite{seurin2025surpassing}. RL has demonstrated superiority over heuristic-based formulations such as Tabu Search, Simulated Annealing, Genetic Algorithms, and Prioritized Replay Evolutionary and Swarm Algorithm \cite{seurin2025surpassing} across various large-scale pressurized-water reactor optimizations. This superiority can be attributed to RL's learning mechanism, which has been delineated in a single-objective RL framework \cite{seurin2024assessment,seurin2025surpassing,seurin2023can}.

Essentially, instead of taking random steps to find better solutions, RL learns which solutions to generate over time, whether based on a scalar or a multi-objective criterion. This results in solutions that improve as the algorithm progresses, rather than exacerbating randomness at all times. Furthermore, the learning process can be accelerated by providing more information to the learning agent, a concept we refer to as ``engineering intuition'' \cite{seurin2024physics}. Specifically, accounting for the underlying correlations between different physical parameters that couple certain constraints with the objectives may lead to designs that are both safer and more economical.

In RL, a policy $\pi(\cdot; \theta)$---where $\theta$ is a learnable parameter---generates solutions as the learning agents traverse the search space. Each agent---the total number of which is determined by the central processing units (CPUs) available for the optimization process---shares the same policy but collects data independently during each rollout. At the end of a rollout, the collected data are aggregated to formulate a loss function, which is then used to update $\theta$. Over time, the policy $\pi(\cdot; \theta)$ is refined to enhance the quality of the solutions generated.
 
 In \cite{seurin2024multiobjective} and \cite{seurin2024physics}, proximal policy optimization (PPO) \cite{schulman2017proximal} was used; however, it should be noted that PPO was chosen here for its simplicity to tune and its performance, but any other policy-based RL algorithm could be utilized and a number of these were tested, including Trust Region Policy Optimization \cite{schulman2017trust}, Asynchronous Advantage Actor-Critic \cite{mnih2016asynchronous}, and Actor Critic using Kronecker-Factored Trust Region \cite{schulman2017trust}.

The PPO loss is a first-order, constraint-free loss function that employs a clipping method to ensure that the policies between updates remain sufficiently close to one another, thereby preventing catastrophic declines in performance. If $r_t(\theta) = \frac{\pi_{\theta}}{\pi_{\theta_{old}}}$ is the probability ratio, the surrogate objective (presented as a loss) to be optimized is the clipped objective:
\begin{equation}
L^{clip}(\theta) = \mathbb{E}(\min(r_t(\theta)A(\theta),\text{clip}(1 - \epsilon,1 +\epsilon, r_t(\theta))A(\theta))),
\label{eq:clipped_obj}
\end{equation}
where $\epsilon$ is a hyperparameter. Moreover, as the advantage estimation $A(\theta)$ was used, another value loss $L^{VF}(\theta) = (V^{\pi_{\theta}}(\theta) - V_{target}(s_t))^2$ was introduced to ensure a proper approximation of the value function (which is used in the Generalized Advantage Estimation methodology to estimate $A(\theta)$ \cite{schulman2017proximal}). Lastly, an entropy term $\mathcal{H}(\theta)$, which corresponds to the entropy of the policy distribution, was introduced to enhance exploration. The final loss becomes:
\begin{equation}
    L^{final}(\theta) = \mathbb{E}(L^{clip}(\theta) - c_{vf}L^{VF}(\theta) + c_{h} \mathcal{H}(\theta)),
    \label{eq:ppoloss}
\end{equation}
where $c_{vf}$ and $c_{h}$ are hyperparameters. 

For this work, we used an augmented objective function that serves as a reward of the form:
\begin{equation}
    f(x) = -\gamma_{lcoe}LCOE + \Phi(\underline{x}),
\end{equation}
where $\gamma_{lcoe}$ is a term associated with the LCOE, equal to 0.1 to balance optimizing the LCOE satisfying the constraints encapsulated in the term $\Phi(\underline{x}) = \sum_{i \in C} \gamma_i \Phi(x_i)$, $\underline{x} = \{x_1,...,x_4\}$. $C = \{ (c_{i})_{i \in \{1,..,4\}} \}$ is the set of constraints, and $(\gamma_{i})_{i \in \{1,..,4\}}$ is a set of weights attributed to each constraint (all equal to 10,000 in this work). If $ (x_{i})_{i \in \{1,..,4\}}$ are the values reached for the core ($x$) for the corresponding constraints $(c_{i})_{i \in \{1,..,3\}}$, $\Phi(x_i) = \delta_{x_i \le c_i} (\frac{x_i - c_i}{c_i})^2$, where $\delta$ is the Kronecker delta function.

\section{Results}
\label{sec:res}

\subsection{Samples Description}
\label{sec:samplesdescription}
We collected 921 samples using the Sawtooth High-Performance Computing Center at Idaho National Laboratory, utilizing one Intel(R) Xeon(R) Platinum 8268 CPU clocked at 2.90 GHz per sample. The computing time for each OpenMC candidate ranged from approximately 40 to 168 hours (the maximum allowable compute time on a Sawtooth node), depending on the fuel lifetime. The QoIs related to neutronics calculations were then extracted, and the fuel lifetime was input into the MOUSE techno-economic assessment tool to obtain the cost performance parameters.

During this process, we also had to discard several outliers that could adversely affect the performance of the surrogate models. These outliers were identified to ensure the remaining samples exhibits reasonable costs and were ``physical.'' Consequently, we removed any instances of negative costs. Ultimately, we retained 875 data points.

We then established constraints for the SDM and fuel lifetime values, set at -6700 pcm (close to that of the nominal design) and 6.0 years (close to the nominal design), respectively. As explained in Section \ref{sec:physicsandtechnoecon}, replacement of the main equipment occurs every 10 years, making it unnecessary to purchase additional fuel to exceed this lifetime. Therefore, we prescribed a 10.40-year upper limit for the fuel lifetime in order to adopt a slightly conservative approach. Additionally, we imposed a constraint on $q^{''}_{max}$ and $F_{\Delta h}$ so as to guide the optimizer toward favorable areas of the search space during objective optimization, whereas no constraints were placed on the LCOE. Of the 875 samples, only two satisfied the constraints, with an associated cost of 11,648 and 8,957 \$/MWh (FOAK). The assignment of constraints and objectives is summarized in Table \ref{tab:assignmentof}.

\begin{table}[htp!]
    \centering
    \caption{Assignment of the objectives and constraints for the optimization with RL. $F_q$ is not utilized here, as it is only input to evaluate the peak heat flux.}
    \begin{tabular}{llllll}
    \hline
    QoIs & Objectives & Constraints & Limits & Sign & Upper/Lower Bound\\
    \hline
  $q^{''}_{max}$ & - & + & 0.020 & decrease & [0.0137, 0.0579]\\ 
  $F_{\Delta h}$ & - & + & 1.47 & increase & [1.164, 1.737]\\
  SDM [pcm] & - & + & -6700 & decrease & [-46526, 8081]\\
  Lifetime & - & + & [6.0,10.40] & increase & [-42.268, 52.434]\\
  \hline
\end{tabular}
    \label{tab:assignmentof}
\end{table}


\subsection{Surrogate Model Performance}
\label{sec:surrogatemodel}

As the surrogate modeling methodologies employed in the current study should not be considered a major contribution of this work, this section is kept intentionally brief to allow for further discussion on other topics. In Subsection \ref{sec:applicationof}, full-order models will be applied to final candidate designs in order to evaluate the accuracy of these surrogate models. 

Optimization of advanced reactors, which do not benefit from established industry-grade legacy reactor physics tools such as SIMULATE \cite{rempe1989simulate}, must rely on more expensive alternatives such as OpenMC \cite{al2023design}. The use of surrogate models is therefore invaluable for accelerating reactor physics computations \cite{seurin2025impact,williams2023novel} and has become increasingly prevalent in the field of nuclear engineering over the past 5 years. For a summary of recent applications, we invite the reader to consult \cite{williams2023novel,che2022machine,seurin2023can}, though it is omitted here for the sake of brevity. For this section, we utilized classical models and compared their performance. As we found the performance to be satisfactory, we opted not to explore more advanced tools at this stage. 

Moreover, we identified that several QoIs (e.g., lifetime, SDM, and $q^{''}_{max}$) are highly correlated; however, only some can be easily predicted based on the input design parameters. Consequently, each QoI will be predicted in two steps. First, we will predict the fuel lifetime, SDM, and $F{\Delta h}$, then use these predictions to augment the input space for predicting the peak heat flux $q^{''}_{max}$. At each step, the best eight inputs are selected for the prediction, using the feature importance option available in the random forest (RF) model from the scikit-learn package \cite{scikit-learn}.

\begin{figure}[htb!]
    \centering
\includegraphics[width=0.8\linewidth]{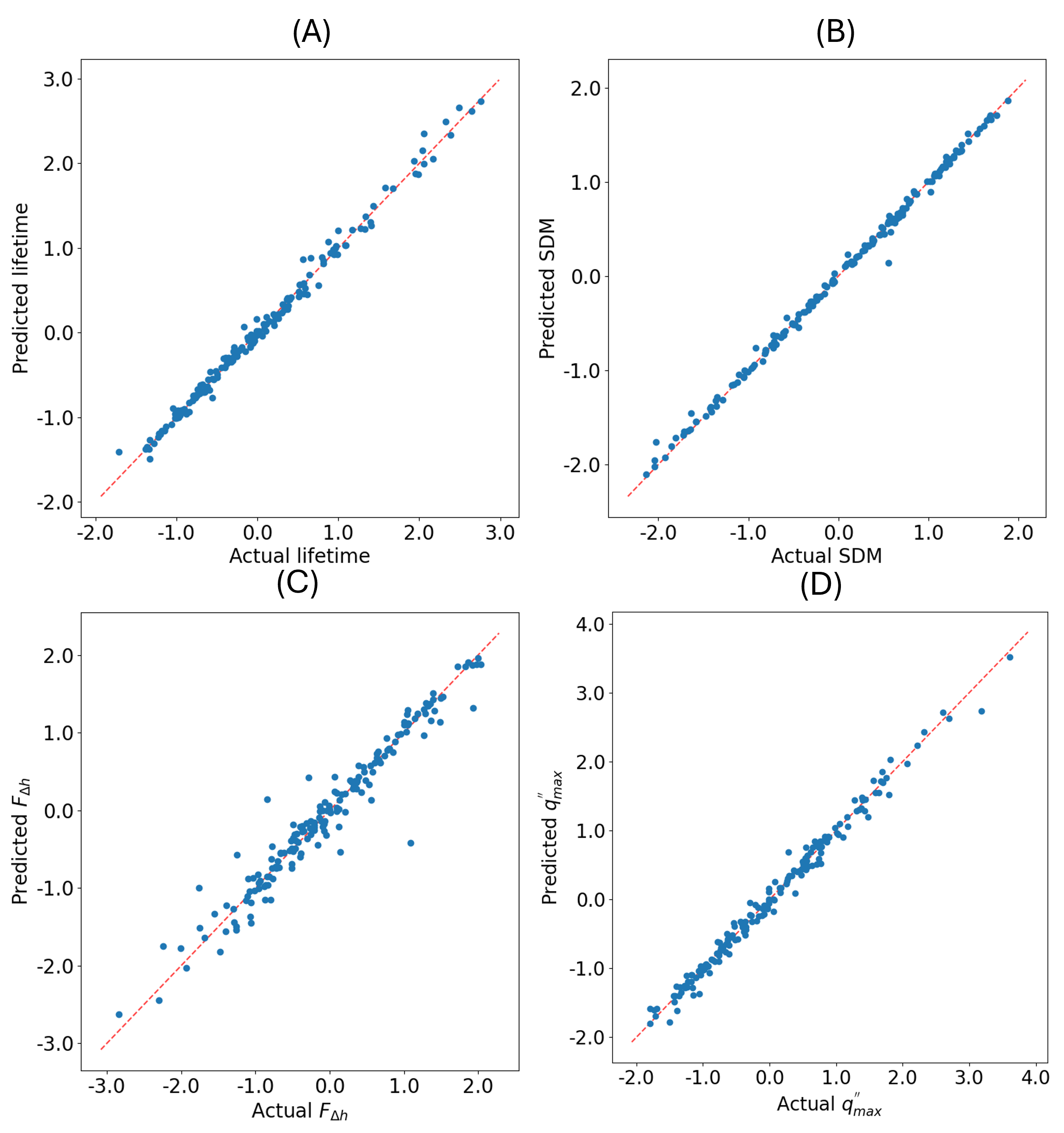}
    \caption{Performance results for the GPs on the first three QoIs: (A) lifetime, (B) SDM, (C) $F_{\Delta h}$, and (F) MLP for $q^{''}_{max}$.}
    \label{fig:gpmlpresults}
\end{figure}

Table \ref{tab:crossvalidationresults} presents the corresponding $R^2$ values from a five-fold cross validation, comparing a linear regression, RF, support vector regression, multi-layer perceptron (MLP), Gaussian process (GP), and eXtreme Gradient Boosting. The highest $R^2$ value is achieved first with the GP model, followed by the MLP for the $q^{''}_{max}$. Consequently, the first batch of QoIs will be predicted using the GP, while the MLP will be employed for predicting $q^{''}_{max}$. Figure \ref{fig:gpmlpresults} illustrates the performance of the surrogate model for each predicted QoI, scalarized using standard normalization $\frac{x_i - \mu_i}{\sigma_i}$, where $\mu_i$ is the mean of the training samples and $\sigma_i$ is the standard deviation of the training samples for the (i)th QoI. For most validation data, the corresponding points lie close to the $y=x$ line, characteristic of a high $R^2$. While the accuracy is not 100\%, it is enough to drive the optimization agents into regions of attraction where the designs found are improving. The full-order model we will compute will provide the real values for the QoIs. 

\begin{table}[htb!]
    \centering
    \caption{Five-fold cross-validation performance of the various models when predicting the QoIs.}
    \begin{tabular}{lllllll}
    \hline
     &  Linear Regression & RF & Support Vector Regression & MLP & GP & eXtreme Gradient Boosting\\
     \hline
    $r^2$ (first batch)     &  0.890 & 0.950 & 0.957 & 0.973 & 0.978 & 0.956\\
    $r^2$ ($q^{''}_{max}$) & 0.713 & 0.897 & 0.957 & 0.979 & 0.955 & 0.920\\
    \hline
    \end{tabular}
\label{tab:crossvalidationresults}
\end{table}

\subsection{Application of the Optimization Methodology, and Analysis of the Optimized Designs}
\label{sec:applicationof}

\begin{figure}[htb!]
    \centering
\includegraphics[width=0.8\linewidth]{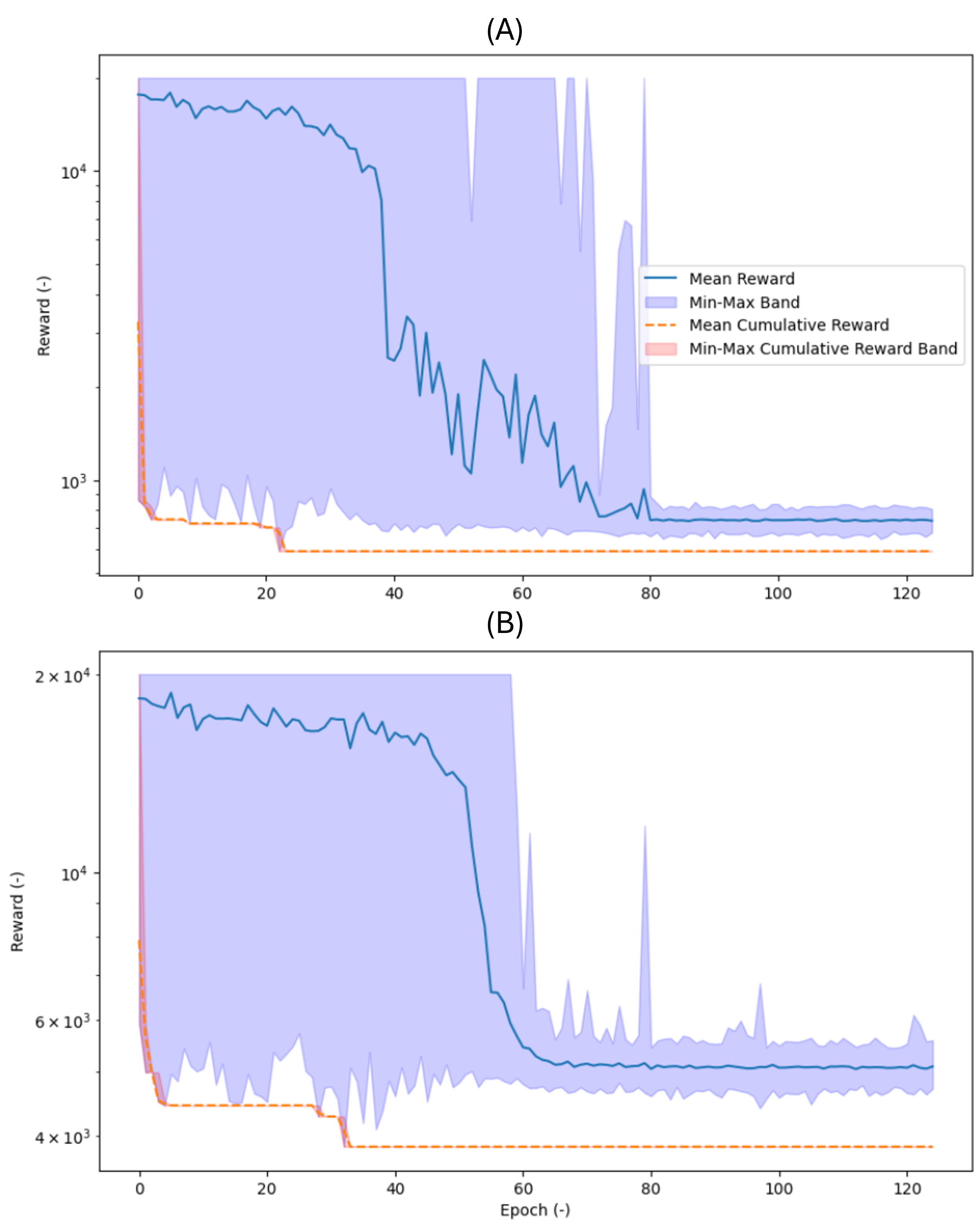}
    \caption{Evolution of the mean and cumulative reward across epochs (1 epoch corresponds to 10,000 samples) for (a) the case of the nominal Be's cost and (B) the replaced-by-graphite cost.}
    \label{fig:pearlresults}
\end{figure}

The PPO-related hyperparameters that influence the optimization (see the stables-baselines3 package \cite{stable-baselines3}, as well as \cite{seurin2024assessment}, for a description of their influence) are n$_{\text{steps}}$ is equal to 8, $c_{\mathcal{H}}$ is 0.0001, $c_{vf}$ is 0.5, the learning rate is 0.00025, the max grad norm is 0.5, the batch size is 0.5, and $\epsilon$ is 0.2. Recall that neither $\gamma$ nor $\lambda$ affect the optimization. We ran the RL algorithm on eight 48 Intel(R) Xeon(R) Platinum 8268 CPUs clocked at 2.90 GHz for 100,000 steps total. The evolutions of the mean and maximum reward, both of which were averaged over chunks of 10,000 samples, are given in Figure \ref{fig:pearlresults}.

The evolutions of the mean and cumulative reward across epochs for the nominal-Be cost case are shown in Figure \ref{fig:pearlresults}(A), and for the graphite case in Figure \ref{fig:pearlresults}(B). For the former, the best solution is found relatively early (within 25 epochs), while the optimizer converges after 80 epochs. However, the uncertainty in the LCOE (FOAK) can be high (350--600 \$/MWh), and the converged solution is within the uncertainty range of the best solution. For the graphite case, the best solution is found after about 35,000 samples, and the distribution of the mean converges after approximately 80,000 samples. Overall, the optimizer behaves as expected, finding feasible solutions and reducing the LCOE.

Furthermore, it is evident that for the nominal-Be cost case, the average cost is significantly higher, due to the very high cost of the Be used in the axial reflector (45,000 \$/kg). However, the optimizer manages to reduce the cost to a similar order of magnitude as the graphite case, though the best solution found is still more expensive than the nominal one (5,844 \$/MWh as compared to the 5,079 \$/MWh evaluated based on the graphite cost). Therefore, we do not anticipate Be ever being the more cost-competitive choice.

Table \ref{tab:inparmsfinal} provides the final input design parameters for the two optimized designs in comparison to the nominal design, and Table \ref{tab:designobjectives} presents the corresponding QoIs.

First, we observe that each new solution exhibits a lower $x_{mr}$. This is notable since we decreased moderation from the solid moderators but increased the reactivity in the core through other pathways. Additionally, as explained in Section \ref{sec:physicsandtechnoecon}, YHx exhibits positive reactivity coefficients. Consequently, even without specifically optimizing for it, we had substantially fewer solid moderator, and thus the ITC ended up being lower for all solutions (see Table \ref{tab:designobjectives}), with -3.04 and -3.00 $pcm/K$ for ``Solution (Be's cost)'' and ``Solution (graphite's cost),'' respectively. Typically, increasing $x_{mr}$ tends to decrease the SDM and $F_{\Delta h}$ (see Figure \ref{fig:correlationmatrix} in \ref{appendix:correlation}); hence, for this specific parameter, the optimization preferentially chose smaller radii, thus benefiting safety as well. Another mechanism was chosen to improve fuel lifetime. In comparing the best solutions for the two different costs, we observe that the main differences lie in $x_{fh}$. For the ``Be's cost'' case, the amount of axial reflector is minimized, requiring more fuel and moderation in order to compensate, resulting in a higher $x_{cr}$ and $x_{pp}$, respectively, as compared to the ``graphite's cost'' case. Naturally, we see in Table \ref{tab:designobjectives} that $q^{''}_{max}$ is higher in the case with lower $x_{fh}$, as it is directly inversely proportional to the amount of fuel in the core. However, in this study, we do not have access to the true fuel performance translation of this value into fuel, monolith, reflector, solid moderator, and HP temperatures, as would be warranted to understand the true impact of a higher-than-nominal $q^{''}_{max}$ (0.0214 $MW/m^2$ for ``Solution graphite's cost'' against 0.0188 for the nominal case). 

\begin{table}[htb!]
\centering
\caption{Parameter descriptions for changing aspects of the candidate designs.}
\label{tab:inparmsfinal}
\begin{tabular}{llll}
\hline
Symbol    &  Solution (Be's cost) & Solution (graphite's cost)        & Nominal                                                                                             \\ \hline
$x_{ca}$  & 91 & 96.01  & 90 \\
$x_{B10}$ & 53 & 62  & 95\\
$x_{fh}$  &  190 & 148.36   & 160\\
$x_{pp}$  &  2.20 & 1.94  & 2.3\\
$x_{e}$   &  0.199 & 0.199  & 0.197\\
$x_{cr}$  & 1.10 & 0.97 & 1\\
$x_{mr}$  & 0.75& 0.689 & 0.825\\
\hline
\end{tabular}
\end{table}

\begin{table}[htp!]
    \centering
    \caption{QoIs associated with the nominal design described in Section \ref{sec:nomdes}, as compared against the best solutions found. The true values are given on the left of each slash, and the predicted ones on the right.}
    \begin{tabular}{llll}
    \hline
    QoIs & Solution (Be's cost) & Solution (graphite's cost)        & Nominal \\
    \hline
      Lifetime   & 11.41/10.24 & 9.61/8.71 & 6.99/8.08\\
                
    SDM & -6708/-7417 & -7952/-7473 & -6757/-6730\\
$F_{\Delta h}$& 1.41/1.40 &1.373/1.343 & 1.469/1.474\\
$q^{''}_{max}$ & 0.016/0.016 & 0.0214/0.0214 & 0.0188/0.0192\\
ITC & -3.04 & -3.00 & -2.404\\
\hline
    \end{tabular}
\label{tab:designobjectives}
\end{table}

Furthermore, Table \ref{tab:designobjectivecost} provides cost-related QoIs for the best designs and the nominal design. For the nominal design, we provide the LCOE values for both the expensive and inexpensive axial reflector scenarios (separated by a slash). In both instances, the fuel is better utilized than in the nominal case. However, when the Be cost is high, the fuel load is significantly higher due to the higher $x_{fh}$ and $x_{cr}$, resulting in a smaller burnup as compared to the case in which the Be cost is low. Ultimately, we see that the LCOE could be decreased by 57\% in the former case, whereas it could be reduced by 78\% if the cost of Be were the same as for the graphite.

\begin{table}[htp!]
    \centering
     \caption{Cost-related QoIs associated with the nominal design described in Section \ref{sec:nomdes}, as compared against the best solutions found. The LCOEs provided for each optimal solution correspond to the cases with associated Be costs. The cost for the nominal case is given with the two costs as in Table \ref{tab:nominaldesignobjectives}.}
    \begin{tabular}{llll}
    \hline
    QoIs & Solution (Be's cost) & Solution (graphite's cost)        & Nominal \\
    \hline
Average Heat Flux & 0.00806 &0.0117 & 0.010536\\
Power Density & 1.612 &2.412 & 2.105\\
Uranium U235 & 149.90 &91.160& 103.44\\
Uranium Mass & 753.27 & 458.09 & 525.06\\
Burnup & 11.07 & 15.33 & 9.725\\
LCOE (FOAK) Estimated Cost (\$2024 USD)& 5844 & 3937 & 10307/5079\\
LCOE (NOAK) Estimated Cost (\$2024 USD)& 1466 & 1292 & 1596/1442\\
\hline
    \end{tabular}
\label{tab:designobjectivecost}
\end{table}

Furthermore, the transportability constraints are respected, as the reactor size remains the same (by design), and the mass of the core components is within the same range (not shown for brevity's sake). Lastly, while the predictions differ slightly, they are still reasonably close, and the surrogate model remains useful for steering decisions within basins of attraction. Future work should include online re-training to improve the performance of the surrogate model as more data are collected.

In the next subsection, we will explore the reasoning behind maximizing or minimizing $x_{fh}$. 

\subsection{Deep Dive into the LCOE}
\label{sec:deepdive}

\begin{figure}[htp!]
    \centering
\includegraphics[width=0.9\linewidth]{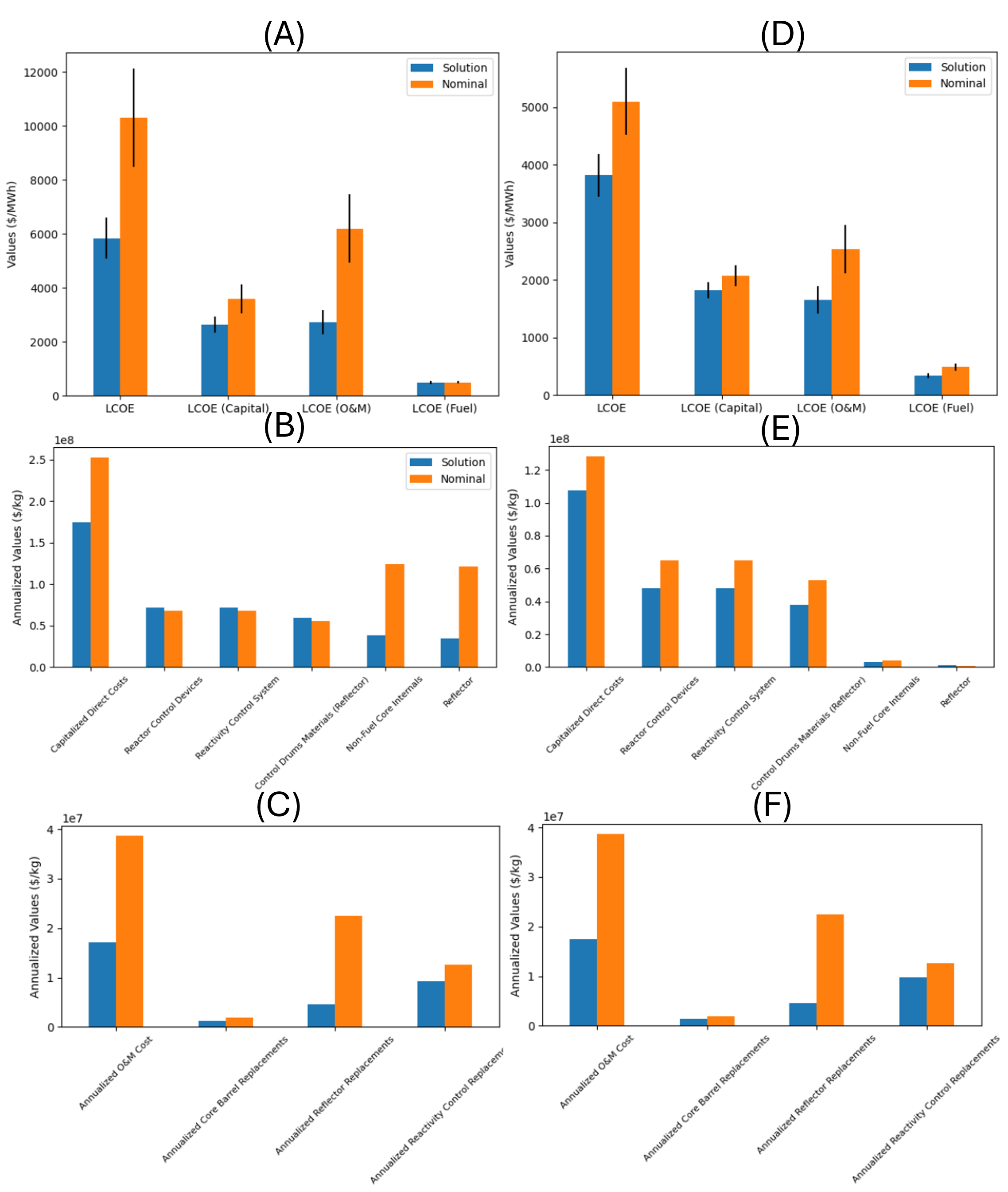}
    \caption{Breakdown of the LCOE per component: fuel cycle cost ((A) and (D)), capital cost ((B) and (E)), and O\&M cost ((C) and (F)).}
    \label{fig:breakdownoflcoe}
\end{figure}

We end our analysis by breaking down the LCOE into its capital, O\&M, and fuel contributions. As shown in Figure \ref{fig:breakdownoflcoe}(A), the capital and O\&M portions dominate the LCOE.  Note that the initial fuel inventory (account 25) is part of the direct capital direct cost and not the annualized fuel cost, but is negligible compared to the other contributions in this work. When we zoom into Figure \ref{fig:breakdownoflcoe}(B) and examine the items composing the annualized capital portion of the LCOE, we see that the reflector and control system costs dominate, with the reflector being the primary contributor. In Figure \ref{fig:breakdownoflcoe}(C), we see that the cost of replacing the reflector and control drums dominates the O\&M portion. This explains why $x_{fh}$ was maximized, as it reduces the impact of the reflector. This resulted in a slightly higher contribution from the control drums, but minimized by a longer fuel lifetime.

Lastly, Figures \ref{fig:breakdownoflcoe}(D)--(F) break down the LCOE and its sub-components for the second case, in which the axial reflector cost is equal to that of graphite. Here, the capital and O\&M portions dominate, but the reflector is inexpensive. As a result, $x_{fh}$ should be minimized to reduce the contribution of the control drum materials, which represent the main contributor to both the capital and O\&M portions of the cost (see Figures \ref{fig:breakdownoflcoe}(E) and \ref{fig:breakdownoflcoe}(F), respectively). However, as we imposed an upper limit for $q^{''}_{max}$, $x_{fh}$ cannot be decreased indefinitely. Expanding the objective space by relaxing this constraint could lower costs even further, and will be the subject of a companion paper.


\section{Conclusion}
\label{sec:conc}
 Nuclear $\mu$R generate less than 20 MWth of power, and thus the U.S.~Department of Energy categorizes them as Category B reactors \cite{naranjo2024assessment}. HPMR stands to be a promising design thanks to its passive heat removal and compactness. However, they are often less economically competitive than alternatives such as gas-cooled $\mu$R, lead-cooled $\mu$R , organically cooled, and water cooled \cite{hanna2025bottom,shirvan2023uo2}. This necessitates that the economy of HPMRs be optimized, and advancement of AI-based optimization methods and surrogate modeling can aid in achieving this goal.

In this work, we leveraged a combination of tools, including OpenMC for neutronics and MOUSE for economic analysis---all wrapped around RL for optimization. Specifically, the cost measured in terms of LCOE for the FOAK was minimized while still respecting the constraints related to the peak heat flux $q^{''}_{max}$, fuel lifetime, SDM, and rod-integrated peaking factor ($F{\Delta h}$). We first generated about 900 samples, then fit the surrogate models used during the optimization, which ran for about 100,000 samples. The best solutions found were evaluated by using a full-order model.

We studied two cases: one in which the axial reflector was very expensive (45,000 \$/kg in \$2024 USD) and one in which it was inexpensive (80 \$/kg in \$2022 USD). First, the solid moderator radius $x_{mr}$ was systematically reduced---as attributed to the associated positive effect on $F_{\Delta h}$---thereby resulting in a lower ITC. We then found that capital and O\&M costs are the main contributors to the LCOE---particularly the cost of the control drum materials and the axial reflectors (when the latter's cost is that of the Be). These contributions originate from both the capitalized direct costs, and the annualized replacement cost incurred by replacing major components including vessels, reflector, moderator, and control drums (every ten years).

When the reflector is expensive, the key strategy is to reduce its cost by increasing $x_{fh}$ and adjusting other parameters so as to ensure operational and safety constraints. On the other hand, an inexpensive axial reflector leads to minimization of $x_{fh}$, which is equivalent to minimizing the amount of control drum materials in the reactor. This increases $q^{''}_{max}$, potentially challenging fuel and HP performance. Ongoing research continues to be focused on integrating fuel performance (using BISON \cite{williamson2012multidimensional}) and HP performance (using Sockeye \cite{hansel2019sockeye}) analyses in order to understand the quantitative impact of such increases and to potentially relax those constraints in hopes of achieving higher cost benefits.

Numerous factors impact the economic performance of $\mu$R, including the core power rating, fuel enrichment, fuel burnup, size of the onsite staff, fabrication costs, policy incentives and financing---as well as the transportability limits, capacity for passive heat removal \cite{buongiorno2021can}, and fuel and HP performance. While the results highly depend on our assumptions (e.g., Be reflector and control drums), our approach offers the advantage of being agnostic to HPMRs, as well as being extendable to any type of reactor, for any type of study of interest. The present work focused on cost, but a companion paper will elucidate the delicate trade-offs between cost and safety when employing multi-objective optimization, thus helping designers make better-informed decisions regarding the future of $\mu$R.

\section*{Credit Authorship Contribution Statement}

 \textbf{Paul Seurin:} Conceptualization, Methodology, Software, Writing – original draft preparation, Funding acquisition, Data curation, Formal analysis \& investigation, Visualization. \textbf{Dean Price:} Conceptualization, Methodology, Software, Writing – original draft preparation, Data curation, Funding acquisition. \textbf{Luis Nunez:} Writing – review \& editing. All authors have read and agreed to the published version of the manuscript.
 
\section*{Acknowledgment}
This work was supported through the Idaho National Laboratory (INL) Laboratory Directed Research \& Development (LDRD) program under U.S.~Department of Energy (DOE) Idaho Operations Office contract no.~DE-AC07-05ID14517. The authors would also like to thank Dr.~Khaldoon Al-Dawood and Dr Botros Hanna for the valuable discussions around the techno-economic analysis of $\mu$Rs, and John Shaver for proof-reading the manuscript.

\section*{Data Availibility}
The data and codes will, by the time of this publication, be made available upon reasonable demand and in an open-source package hosted on the \href{https://github.com/IdahoLabResearch}{IdahoLabResearch GitHub repository}.

\appendix
\section{Correlation matrix connecting design parameters and QoIs}
\label{appendix:correlation}

\begin{figure}[htp!]
     \centering
\includegraphics[width=1.0\linewidth]{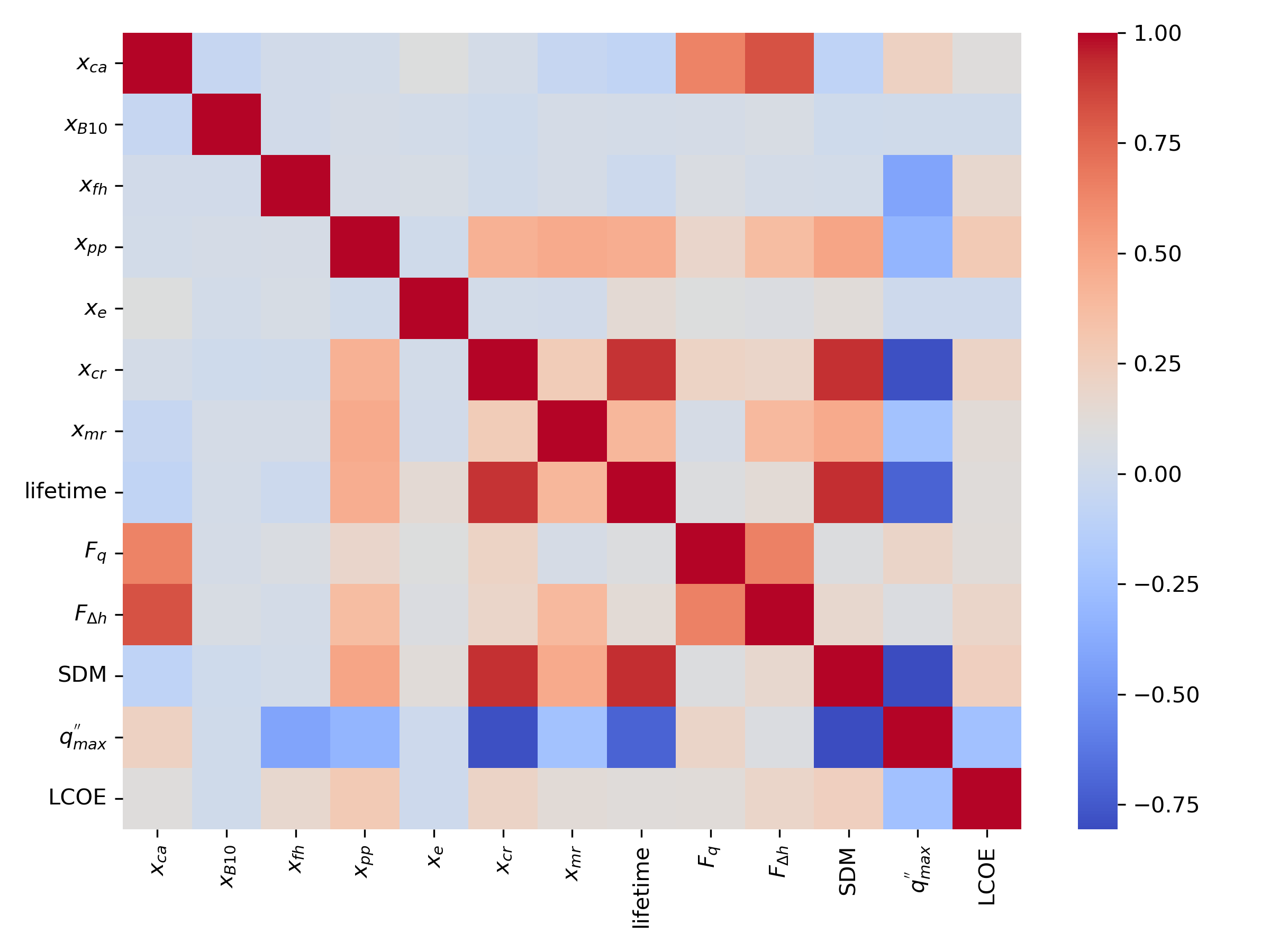}
     \caption{Correlation matrix between the input design parameters and the QoIs. ``1'' signifies a strong positive linear correlation and ``-1'' a strong negative one.}
     \label{fig:correlationmatrix}
 \end{figure}

Understanding the correlation between the input design parameters and the QoIs is crucial for trusting and interpreting the design decisions made. This relationship is typically elucidated using correlation matrices \cite{seurin2024physics,kobayashi2024explainable}. While various methods exist, including SHAP, LIME, and MRMR \cite{kobayashi2024explainable}, this appendix presents the (linear) correlation matrix between the input design parameters and the QoIs (see Figure \ref{fig:correlationmatrix}). This approach was deemed sufficient for the purposes of this study.

\bibliographystyle{elsarticle-num-names} 
\bibliography{refs.bib}

\end{document}